\documentclass[conference]{IEEEtran}

\usepackage{times}
\usepackage{latexsym}
\usepackage{graphicx}
\usepackage{mathtools}
\usepackage{multirow}
\usepackage{euler}
\usepackage{array}
\usepackage{pgfplots}
\usepackage{pgfplotstable}
\usepackage{pgf-pie}
\usepackage{color, colortbl}

\begin{document}
%
\title{Augmenting Visual Question Answering with Semantic Frame Information in a Multitask Learning Approach}

\author{\IEEEauthorblockN{Mehrdad Alizadeh}
\IEEEauthorblockA{University of Illinois at Chicago\\
Chicago, Illinois, USA\\
Email: maliza2@uic.edu}
\and
\IEEEauthorblockN{Barbara Di Eugenio}
\IEEEauthorblockA{University of Illinois at Chicago\\
Chicago, Illinois, USA\\
Email: bdieugen@uic.edu}}

\maketitle

\begin{abstract}
Visual Question Answering (VQA) concerns providing answers to Natural Language questions about images. Several deep neural network approaches have been proposed to model the task in an end-to-end fashion. Whereas the task is grounded in visual processing, if the question focuses on events described by verbs, the language understanding component becomes crucial. Our hypothesis is that models should be aware of verb semantics, as expressed via semantic role labels, argument types, and/or frame elements. Unfortunately, no VQA dataset exists that includes verb semantic information. Our first contribution is a new VQA dataset (imSituVQA) that we built by taking advantage of the imSitu annotations. The imSitu dataset consists of images manually labeled with semantic frame elements, mostly taken from FrameNet. Second, we propose a multitask CNN-LSTM VQA model that learns to classify the answers as well as the semantic frame elements. Our experiments show that semantic frame element classification helps the VQA system avoid inconsistent responses and improves performance.
\end{abstract}

\IEEEpeerreviewmaketitle

\section{Introduction} \label{introduction}
The goal of a Visual Question Answering (VQA) system is to answer user questions about an image \cite{VQA}. In order to train neural-network based VQA models, many large-scale datasets have been created \cite{kafle2017visual}. We have observed that a large portion of the questions available in current datasets, involve a verb other than "\textit{to be}" (i.e. $42$\% of VQA dataset). Questions including "\textit{to be}" as the primary verb are usually about \textit{objects}, \textit{object attributes}, \textit{object presence}, \textit{object frequency}, \textit{spatial reasoning} and so on. These questions appear to be more visually than linguistically challenging.
On the other hand,  event verbs such as \textit{cook} or {\it jump}, inherently provide semantic information that may help in answering questions about images describing such events.  Semantic information about verbs includes the type of arguments a verb can take and how the arguments participate in the event expressed by a verb, but this information is missing in current VQA systems. We contend that, if a VQA system is aware of such semantic information,  it can not only narrow down the possible answers but also avoid providing irrelevant responses. For example, the answer to the question "\textit{What is the woman cooking in the oven?}", should belong to the  \textit{food} semantic category. However, neither do VQA datasets encode, nor has any  VQA system taken advantage of this information. \\
The question is how to incorporate such semantic information in VQA. Traditionally in linguistics, semantic  information about a verb has been captured via so-called {\it thematic} or {\it semantic roles} \cite{martin2009speech}, which may include roles like {\it agent} or {\it patient} as encoded in a resource such as VerbNet \cite{kipper08}.
Semantic role labeling has been shown to improve performance in challenging tasks such as dialog systems, machine reading, translation and question answering \cite{strubell2018linguistically,shen2007using}.  However, the difficulty of clearly defining such roles has given rise to other approaches, such as the abstract roles provided by PropBank \cite{propbank05}, or the specialized frame elements provided by FrameNet \cite{fillmore2003background}. In FrameNet, verb semantics is described by frames or situations.  Frame elements are defined for each frame and correspond to major entities present in the evoked situation. For example, the frame {\it Cooking\textunderscore creation} has four core elements, namely {\it  Produced\textunderscore food}, {\it Ingredients,} {\it Heating\textunderscore Instrument}, {\it Container}. \\
In order to create a VQA dataset with verb semantic information, 
we took advantage of the imSitu dataset \cite{yatskar2016},
developed for situation recognition and consisting of about $125k$ images. Each image is annotated with one of  $504$ candidate verbs and its frame elements according to FrameNet \cite{fillmore2003background}. A sample of images from the ImSitu dataset and their annotations can be found in Table~\ref{images-table}.\footnote{imSitu substitutes some frame elements with more traditional thematic roles, for example {\it Agent} for {\it Cook}.}

\definecolor{LightCyan}{rgb}{0.95,0.95,1}
\begin{table*}[h!]
\begin{center}
\setlength{\tabcolsep}{5pt}
\renewcommand{\arraystretch}{0.75}
\begin{tabular}{>{\footnotesize}r >{\footnotesize}l >{\footnotesize}r >{\footnotesize}l >{\footnotesize}r >{\footnotesize}l>{\footnotesize}r >{\footnotesize}l}
\multicolumn{8}{c}{\textbf{}}\\
\rowcolor{LightCyan}
\multicolumn{4}{c}{\textbf{cooking}} & \multicolumn{4}{c}{\textbf{buying}}\\
&&&&&&&\\
\multicolumn{2}{c}{\includegraphics[width=3cm, height=3cm]{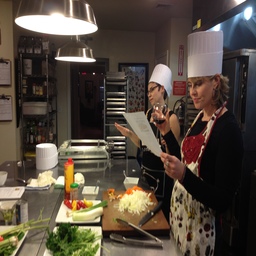}}& 
\multicolumn{2}{c}{\includegraphics[width=3cm, height=3cm]{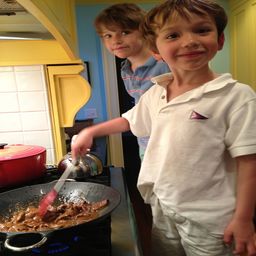}}&
\multicolumn{2}{c}{\includegraphics[width=3cm, height=3cm]{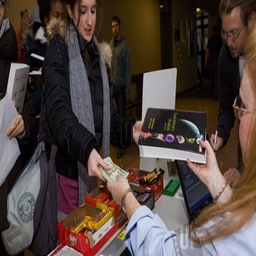}}&
\multicolumn{2}{c}{\includegraphics[width=3cm, height=3cm]{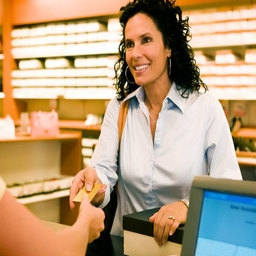}}\\

Agent & woman & Agent & boy & Agent & adolescent & Agent & woman \\
Food & vegetable&Food &meat &Goods  & book & Goods & shoe  \\
Container &pot&Container &wok&Payment & cash & Payment & credit card\\
Tool & knife&Tool &spatula &Seller &  & Seller & person  \\
Place & kitchen& Place & kitchen&Place &  & Place & shoe shop \\
 & & & & & \\
 \rowcolor{LightCyan}
 \multicolumn{4}{c}{\textbf{catching}} & \multicolumn{4}{c}{\textbf{opening}}\\
&&&&&&&\\
\multicolumn{2}{c}{\includegraphics[width=3cm, height=3cm]{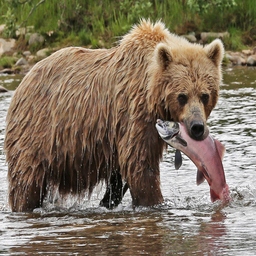}}& 
\multicolumn{2}{c}{\includegraphics[width=3cm, height=3cm]{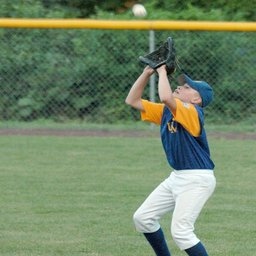}}&
\multicolumn{2}{c}{\includegraphics[width=3cm, height=3cm]{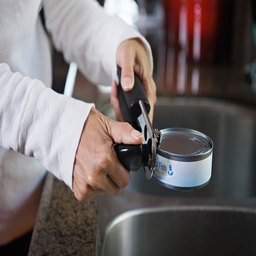}}&
\multicolumn{2}{c}{\includegraphics[width=3cm, height=3cm]{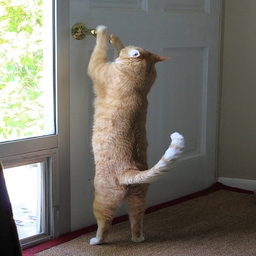}}\\
Agent&bear&Agent&ballplayer&Agent&person&Agent&cat\\
Caughtitem&fish&Caughtitem&baseball&Item&can&Item&door\\
Tool&mouth&Tool&baseball glove	&Tool&can opener&Tool&paw\\
Place&body of water&Place&outdoors&&&&\\

\end{tabular}
\end{center}
\caption{\label{images-table} Sample imSitu annotations of images about cooking, buying, catching and opening. \cite{yatskar2016}}
\end{table*}

In this paper, we first show how we created the new imSituVQA dataset, by employing a semi-automatic approach to create question-answer pairs derived from the imSitu dataset. We have recently publicly released the  imSituVQA dataset\footnote{https://github.com/givenbysun/imSituVQA}. Our second contribution is an augmented CNN-LSTM VQA model with semantic frame element information in a multi-task learning paradigm. The model is trained to classify answers as well as semantic frame elements. The two classifiers share the same weights and architectures up to the classification point. The experiments show that the frame element classification acts like a regularizer by reducing the inconsistencies between the two members of the predicted $<$\textit{answer, frame element}$>$ pair in order to provide accurate responses.

\section{Related Work}
\label{relatedwork}

\subsection{VQA Datasets}
\label{relatedwork_dataset}
In order to train neural-network based VQA models, many large-scale datasets have been created. Datasets differ based on the number of images, the number of questions, complexity of the questions, reasoning required and content information included in the annotation for images, and questions. AQUAR  \cite{malinowski2014nips} is among the first benchmarks released for the VQA task. It includes visual questions on color, number and physical location of an object. In the COCO-QA dataset \cite{ren2015exploring} questions are generated from image captions describing the image. The VQA dataset \cite{VQA}, among widely used benchmarks, is a collection of diverse free form open ended questions. Visual7w \cite{zhu2016visual7w} is a dataset with the goal of providing semantic links between textual descriptions and image regions by means of object-level grounding. FVQA \cite{wang2017fvqa} primarily contains questions that require external information to answer. \\

\subsection{VQA Methods}
\label{relatedwork_methods}

Numerous baselines and methods have been proposed for the VQA task. The VQA task requires co-reasoning over both image and text to infer the correct answer. Most existing methods formulate VQA as a classification problem and impose the restriction that the answer can only be drawn from a fixed answer space. The current dominant baseline method proposed in \cite{VQA} employs a CNN-LSTM-based architecture. It consists of a convolutional neural network (CNN) to extract image features and a long short term memory network (LSTM) to encode the question features. The method fuses these two feature vectors via an element-wise multiplication and then passes the result vector through fully connected layers to generate a softmax distribution over output answers. \\
The attention techniques learn to focus on the most discriminative regions rather than the whole image to guide the reasoning for finding the answer. Different attention techniques, such as stacked attention \cite{yang2016stacked}, co-attention between question and image \cite{lu2016hierarchical}, and factorized bilinear pooling with co-attention \cite{yu2017multi} have been shown to improve the performance of  VQA.

\section{The imSituVQA Dataset}

\begin{table*}[h!]
\begin{center}
\resizebox{0.6\textwidth}{!}{%
\begin{tabular}{|c|c|c|} 
 \hline
\textbf{Verb} & \textbf{Question} \textbf{Template} & \textbf{Frame Element} \\
 \hline 
  & Who is cooking? & AGENT\\ 
 & What does the AGENT cook with TOOL? & FOOD\\
cooking  & What is the AGENT doing?  & VERB\\ 
 & What does the AGENT use to cook  in CONTAINER? & TOOL \\
 & Where does the AGENT cook FOOD  in CONTAINER? &  PLACE\\
 \hline
 & Who is buying GOODS? & AGENT \\
 & What is the AGENT doing? & VERB \\
buying &  What item does the AGENT buy with PAYMENT? & GOODS \\
 & Who does the AGENT buy GOODS from? & SELLER \\
 & Where does the AGENT buy GOODS? & PLACE \\
 \hline
 & Who catches at PLACE? & AGENT\\
 catching & What is the AGENT doing? & VERB \\
 & What item does the AGENT catch with TOOL? & CAUGHTITEM \\ 
 & Where does the AGENT catch CAUGHTITEM? & PLACE\\ 
 \hline
 & What does the AGENT use to open ITEM? & TOOL \\ 
 opening & Who opens ITEM? & AGENT \\
 & What item does the AGENT open? & ITEM \\
 & Where does the AGENT open ITEM with TOOL? & PLACE\\
 \hline
\end{tabular}}%
\end{center}
\caption{\label{questions-table} A subset of Question Answer templates generated for cooking, buying, catching and opening.  }
\end{table*}

In this section, we first briefly expand on our earlier description of imSitu, and explain how question-answer templates are generated. We then describe how they are filled with noun values from the imSitu annotated images. The process results in the creation of a new dataset, which we call imSituVQA. 
As we noted, the imSitu dataset \cite{yatskar2016} is tailored to situation recognition, a problem that involves predicting activities along with actors, objects, substances, and locations and how they fit together. imSitu utilizes linguistic resources such as FrameNet and WordNet in order to define a  comprehensive space of situations. It provides representations helping to understand  who (\textit{AGENT}) did what (\textit{ACTIVITY}) to whom (\textit{PATIENT}), where (\textit{PLACE}), using what (\textit{TOOL}) and so on. \\
Every situation in imSitu is described with one of $504$ candidate verbs such as \textit{cook}, \textit{play}, \textit{tattoo}, \textit{wash}, \textit{teach} and so on. Each verb has a set of FrameNet related frame elements\footnote{As noted earlier, some are traditional thematic roles such as \textit{AGENT} and not the corresponding FrameNet frame elements.}:  for example, $S_{r}($\textit{cooking}$)$ =$\{$ \textit{AGENT, FOOD, CONTAINER, HEATSOURCE, TOOL, PLACE }$\}$ indicates semantic frame elements of the verb \textit{cooking}. This set is also expressed by an abstract definition: \textit{"an AGENT cooks a FOOD in a CONTAINER over a HEATSOURCE using a TOOL in a PLACE"}. imSitu includes $190$ unique frame elements, some shared among verbs such as \textit{AGENT} and some verb specific such as \textit{PICKED}$\in S_{r}(\textit{PICKING})$. \\
Every image is labeled with one of $504$ candidate verbs along with frame elements filled with noun values from WordNet. If an element is not present in the image its value is empty. There are about $250$ images per verb and $3.55$ frame elements per verb on average. 

\subsection{Question answer template generation}
Before template generation, we mapped every frame element to a question word, for example, \textit{AGENT} to \textit{who}, \textit{LOCATION} to \textit{where},  \textit{ITEM} , \textit{FOOD} and \textit{PICKED} to \textit{what item}, \textit{TOOL} to \textit{what  does [AGENT] use to} and so on. From $190$ unique frame elements, $47$ were mapped to \textit{who}, $19$ mapped to \textit{where}, $53$ mapped to \textit{what}  and the remaining were mapped to a question word starting with \textit{what} such as \textit{what item}.\\
There are $504$ abstract definitions, each expressing a verb with its frame elements in a sentence. Given an abstract definition, we hold out one element as output frame element and use the remaining ones in order to generate question templates. For example, for \textit{cook} the abstract definition is \textit{"an AGENT cooks a FOOD in a CONTAINER over a HEATSOURCE using a TOOL in a PLACE"}. If we hold out \textit{FOOD} then what remains is \textit{"an AGENT cooks [X] in a CONTAINER over a HEATSOURCE using a TOOL in a PLACE"}. We created a recursive template question generation procedure that produces all possible combinations. For example, asking about \textit{FOOD} requires templates staring with the \textit{"What ..."} question word, then including or excluding other frame elements in the question results in different possible questions: \textit{"What does AGENT cook?"}, \textit{"What does AGENT cook with TOOL?"}, \textit{"What does AGENT cook in CONTAINER?"} and so on. One advantage of this process is to generate many training samples useful for training deep models. 
A subset of templates for \textit{cooking}, \textit{buying}, \textit{catching} and  \textit{opening} are shown in Table~\ref{questions-table}. The abstract definitions for \textit{buying}, \textit{catching} and  \textit{opening} are:  \textit{"AGENT buys GOODS with PAYMENT from the SELLER in a PLACE" },
\textit{"an AGENT catches a CAUGHTITEM with a TOOL at a PLACE"} and 
\textit{"the AGENT opens the ITEM with the TOOL at the PLACE"}.
In total, $6879$ templates are generated, with on average $13.65$ question-answer templates per verb.

\definecolor{LightCyan}{rgb}{1,1,1}
\begin{table*}[h!]
\begin{center}
\setlength{\tabcolsep}{5pt}
\renewcommand{\arraystretch}{0.75}
\begin{tabular}{>{\footnotesize}c >{\footnotesize}c c >{\footnotesize}c >{\footnotesize}c >{\footnotesize}c}

\rowcolor{LightCyan}
\multicolumn{3}{c}{\textbf{IMAGE} about cooking} &
\multicolumn{3}{c}{\textbf{IMAGE} about buying} \\
&&&&&\\
\multicolumn{3}{c}{\includegraphics[width=4cm, height=4cm]{figures/cooking_samples/cooking_21.jpg}} &
\multicolumn{3}{c}{\includegraphics[width=4cm, height=4cm]{figures/buying_samples/buying_7.jpg}}  \\ 
&&&&&\\
 \textbf{QUESTION} & \textbf{ANSWER} & \textbf{FRAME} & \textbf{QUESTION} & \textbf{ANSWER} & \textbf{FRAME}\\ 
 & & \textbf{ELEMENT} & & & \textbf{ELEMENT}\\
Who is cooking? & boy & AGENT & Who is buying shoes? & woman&AGENT\\ 
What does the boy cook with spatula? & meat & FOOD & What is the woman doing? & buying & VERB\\
What is the boy doing?  & cooking  & VERB & What item does the woman buy & shoe & GOODS\\ 
What does the boy use to cook in wok? & spatula  & TOOL&with credit card? & &\\
Where does the boy cook meat in wok? & kitchen & PLACE & who does the woman buy shoe from? & person& SELLER\\
& & & where does the woman buy shoe? & shoe store & PLACE\\
&&&&&\\
\rowcolor{LightCyan}
\multicolumn{3}{c}{\textbf{IMAGE} about catching} &
\multicolumn{3}{c}{\textbf{IMAGE} about opening} \\
&&&&&\\
\multicolumn{3}{c}{\includegraphics[width=4cm, height=4cm]{figures/catching_samples/catching_24.jpg}} &
\multicolumn{3}{c}{\includegraphics[width=4cm, height=4cm]{figures/opening_samples/opening_251.jpg}}  \\ 
&&&&&\\
 \textbf{QUESTION} & \textbf{ANSWER} & \textbf{FRAME} & \textbf{QUESTION} & \textbf{ANSWER} & \textbf{FRAME}\\ 
 & & \textbf{ELEMENT} & & & \textbf{ELEMENT}\\
who catches at body of water? & bear & AGENT & what does the cat use to open the door? & paw & TOOL\\ 
what is the bear doing? & catching & VERB & who opens the door? & cat & AGENT\\
where does the bear catch fish?  & body of water  & PLACE &  what item does the cat open? & door & ITEM\\ 
what item does the bear catch? & fish & CAUGHTITEM
&	&\\
with mouth? & & & & \\
\end{tabular}
\end{center}
\caption{\label{imSituVQA-samples-table} imSituVQA dataset samples about cooking, buying, catching and opening. }
\end{table*}

\subsubsection{Question answer pair realization}
\begin{center}
\begin{figure}[h!]
\begin{center}
\includegraphics[width=0.35\textwidth]{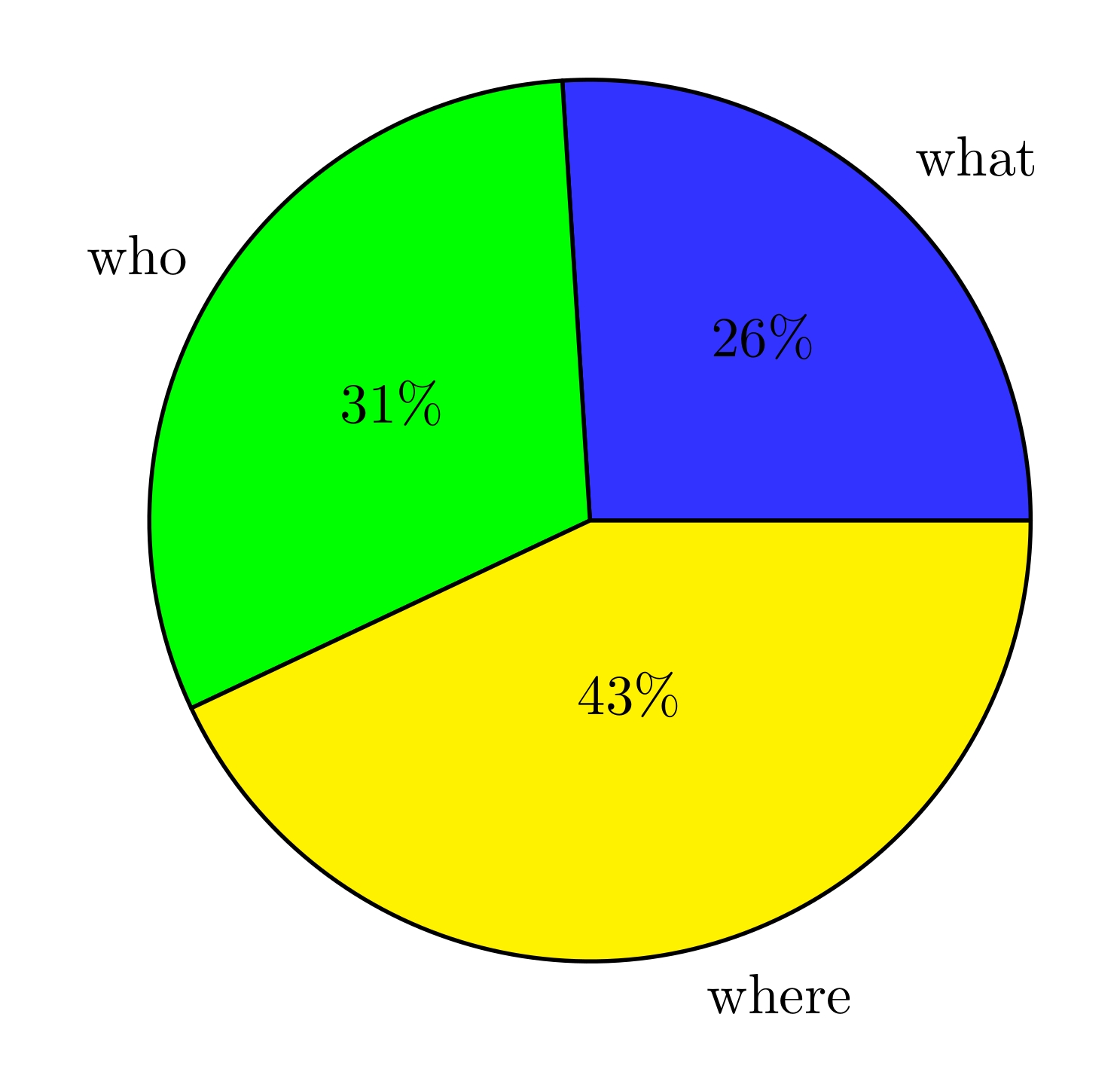}
\caption{Distribution of questions in imSituVQA.}
\label{fig:imsituvqadisttype}
\end{center}
\end{figure}
\end{center}

\begin{center}
\begin{figure}[h!]
\includegraphics[width=0.5\textwidth]{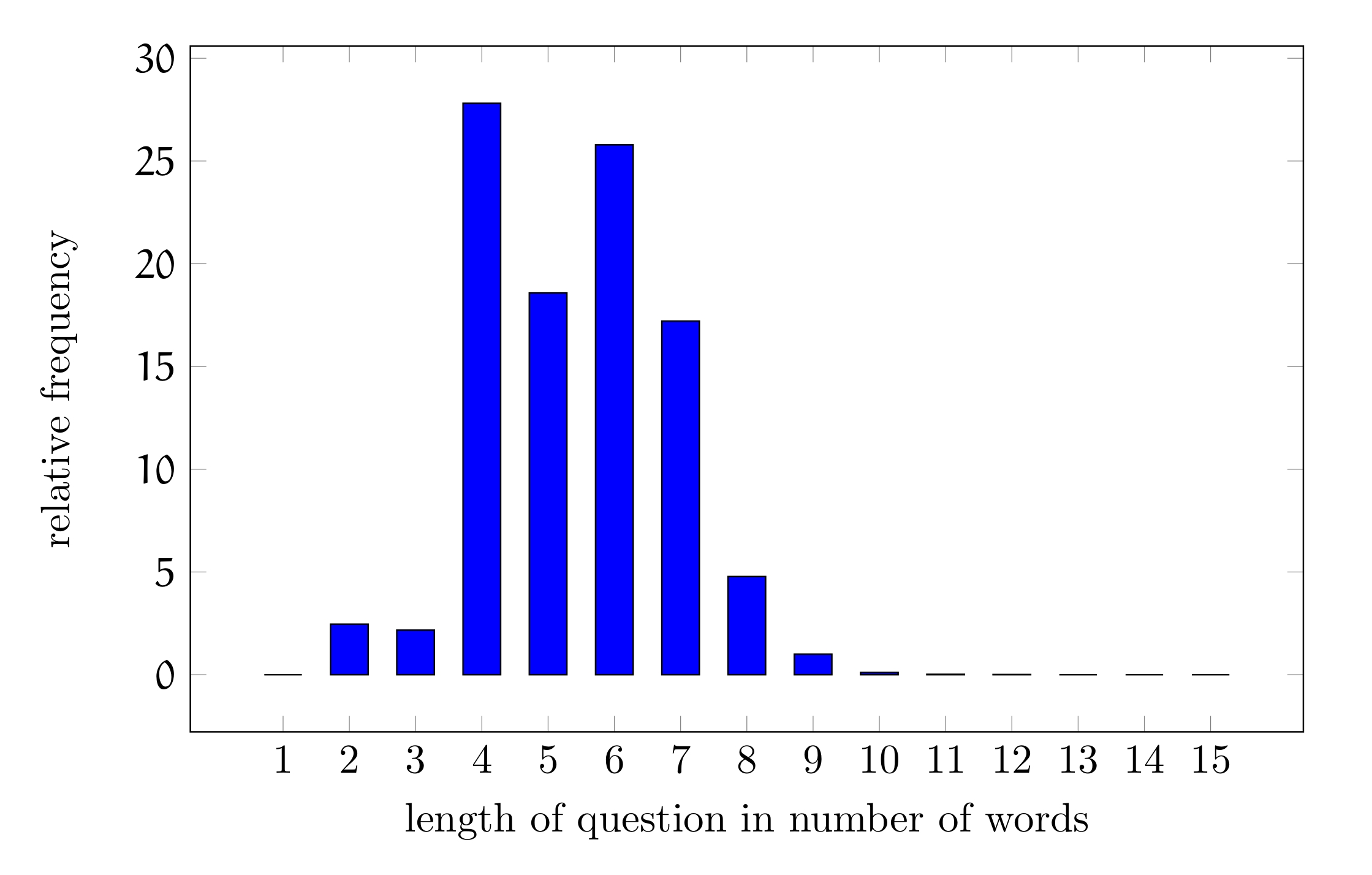}
\caption{Distribution of questions in imSituVQA based on length.}
\label{fig:imsituvqadistlen}
\end{figure}
\end{center}

The template generation is based on $504$ abstract definitions of the verbs. In order to build the real imSituVQA dataset, image annotations are used to substitute the frame elements in the templates. Each image is annotated with a verb and its frame elements with their fillers. Table~\ref{images-table} shows an example of such annotations for \textit{cooking}, \textit{buying}, \textit{catching} and \textit{opening}. All templates of a verb can be instantiated by filling frame elements with noun values from the annotation. If a verb has $n$ templates, applying an image annotation results in $n$ real \textit{$<$question, answer$>$} samples of the image.  Table~\ref{imSituVQA-samples-table} shows VQA samples for \textit{cooking}, \textit{buying}, \textit{catching} and \textit{opening}. This way, the size of the extracted dataset is the average number of templates times the number of images. This realization process results in $254k$ train, $88k$ development and $88k$ test samples. For the training set, the top 10 most frequent frame element classes among the existing  $190$ are shown in Table ~\ref{frequent-table2}. Table ~\ref{frequent-table1} also shows the top $10$ frequent answers. Because $60\%$ of answers are about \textit{place} and \textit{agent}, the most frequent answers are usually values from these two frame elements. Figure~\ref{fig:imsituvqadistlen} depicts the distribution of question template lengths in terms of the number of words. The questions are mostly between 4 to 7 words. Figure~\ref{fig:imsituvqadisttype} shows the distribution of imSituVQA questions according to the first question word. As can be seen \textit{"Where"} is more frequent than \textit{"Who"} and \textit{"What"}. This derives from \textit{place} being the most frequent frame element, twice as frequent as \textit{agent}, which is the second. 

\begin{table}[h!]
\begin{center}
\begin{tabular}{c c} 
 \hline
Frame element & frequency \\
\hline 
\textit{PLACE} & 100,006  \\
\textit{AGENT} & 49,976 \\
\textit{ITEM} & 24,376  \\
\textit{TOOL} & 13,908  \\
\textit{VICTIM} & 3,932 \\
\textit{TARGET} & 3,860 \\
\textit{VEHICLE} & 3,706 \\
\textit{DESTINATION} & 3,238 \\
\textit{COAGENT} & 2,544 \\
\textit{OBJECT} & 2,317 \\
\hline 
\end{tabular}
\end{center}
\caption{\label{frequent-table2} Top 10 frequent frame elements  in imSituVQA training samples. }
\end{table}

\begin{table}[h!]
\begin{center}
\begin{tabular}{c c} 
 \hline
Answer &frequency\\
\hline 
outdoors & 14,621 \\
man & 13,527 \\
woman & 10,763 \\
people & 9,228 \\
room & 8,323 \\
outside & 6,881 \\
inside & 6,679 \\
person & 5,625 \\
hand & 4,238 \\
field & 3,086 \\
\hline 
\end{tabular}
\end{center}
\caption{\label{frequent-table1} Top 10 frequent answers in imSituVQA training samples. }
\end{table}

\begin{figure*}[h!]
\includegraphics[width=\linewidth]{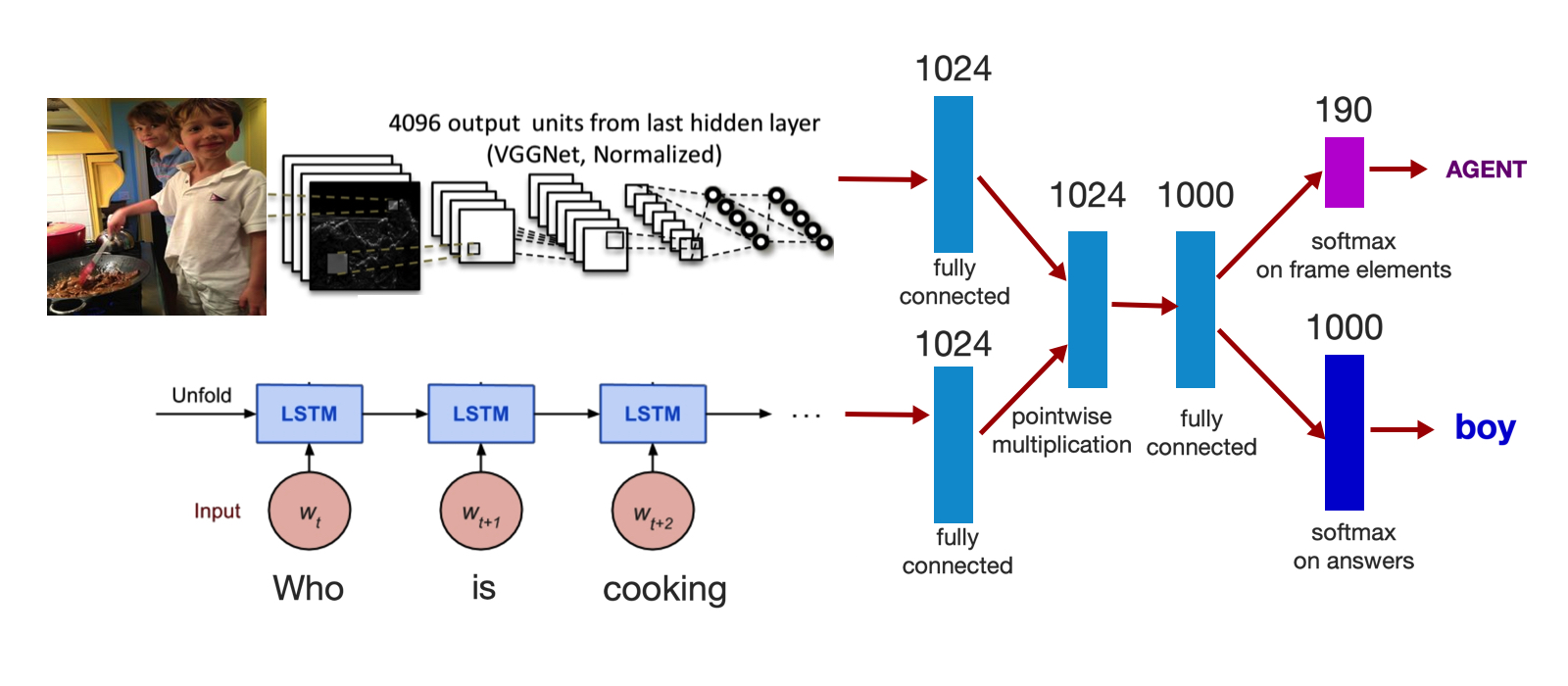}
\caption{Proposed multitask learning architecture for VQA}
\label{fig:boat1}
\end{figure*}

\begin{table*}[h!]
\begin{center}
\begin{tabular}{c c c} 
\hline
 & Accuracy (\%) & WUPS at 0.9 (\%)\\
\hline 
prior ("outdoors") & 05.68  & 11.87\\
per verb prior & 22.15  & 27.65\\
CNN-LSTM  & 39.58  & 46.92\\
multi-task CNN-LSTM  & \textbf{44.90} & \textbf{51.83}\\
\hline 

\end{tabular}
\end{center}
\caption{\label{perfromance-table} Performance of our VQA model on imSituVQA dataset }
\end{table*}

\section{Our VQA Model}
Hyper-class augmented deep learning model has been shown to work well for fine-grained image classification. Instead of fine tuning a convolutional neural network (CNN)  , \cite{xie2015hyper} suggests a hyper-class augmentation formulated as multi-task learning in order to boost the recognition task.   Similarly, we include frame element classification in parallel with answer classification. \\
Let $D_{t} = \{(x_{1}^t , y_{1}^t ), ... , (x_{n}^t , y_{n}^t )\}$ be a set of training \textit{$<$image, question$>$} paired samples with $y_{i}^t \in \{1, ... , C\}$ indicating the answers (e.g., \textit{child}, \textit{kitchen} and \textit{cooking}) of \textit{$<$image, question$>$} pair $x_{i}^t$ , and let $D_{a} = \{(x_{1}^a , r_{1}^a ), ... , (x_{n}^a , r_{n}^a)\}$ be a set of auxiliary frame element information, where $r_{i} \in \{1, ... , R\}$ indicates the frame element class of \textit{$<$image, question$>$} pair $x_{a}^t$ (e.g., \textit{AGENT}, \textit{FOOD} and \textit{PLACE}). 
The goal is to learn a VQA model that correctly answers to an input \textit{$<$image, question$>$} pair. In particular, we
aim to learn a prediction function given by $Pr(y|x)$, i.e., given the input \textit{x:$<$image, question$>$} pair, we compute the probability that $y$ is the answer. Similarly, we let $Pr(r|x)$ denote the frame element classification model. Given the training \textit{$<$image, question$>$} pairs and the answers with auxiliary frame element information, our strategy is to train a multi-task deep model. This model uses a shared CNN-LSTM VQA architecture up to the classification layer. Then sharing common features, it branches out to two different classifiers. One classifier classifies answers,  and the other one, frame elements. Figure ~\ref{fig:boat1} summarizes the proposed multi-task learning model. In order to train the proposed VQA model, the total loss is the average of losses from these two classifiers.   

\section{Evaluation}

\subsection{Experimental Setup}
The proposed VQA model is evaluated by means of the CNN-LSTM-based architecture introduced in \cite{antol2015vqa}. Training deep models requires significant time and resources. Consequently, we employ  trained models such as GLOVE \cite{pennington2014glove} and VGG-NET \cite{simonyan2014very}. GLOVE provides a good word embedding layer initialization that generalizes well and provides a performance boost. GLOVE 300-dimensional weights are utilized in order to feed question words to a bidirectional long short term memory network (LSTM). The output of the LSTM is a 300 dimension question embedding which is mapped to 1024 dimensions by passing through a nonlinear layer. A VGG-NET-16 pre-trained model was used in order to extract image feature vectors. The 4096 image embedding is mapped to 1024 dimensions by passing through a nonlinear layer. The multimodal fusion of image and question embeddings occurs via pointwise multiplication, then after passing through two nonlinear layers with \textit{tanh} activation function, the final embedding is fed to the frame element softmax layer and the answer softmax layer. The model is trained by minimizing the sum of the two cross-entropy loss functions using the rmsprop optimization algorithm \cite{tieleman2012lecture}. The training data is passed with a batch size of $500$ in $50$ epochs. \\

\subsection{Results and Discussions}
Table~\ref{perfromance-table} shows the performance evaluation of the test samples. Using the most frequent answer (prior) in order to answer each question results in $5.65\%$ accuracy. Selecting the most frequent answer per verb results in $22.15\%$ accuracy. The CNN-LSTM model was trained with single answer softmax ($39.58\%$ accuracy) and multi-task, including both answer softmax and frame element softmax ($44.90\%$ accuracy). Augmenting VQA with frame element information boosts the accuracy up to $5\%$. This improvement in the generalization of the CNN-LSTM model indicates how well the multi-task approach acts like a regularizer. A chi-square test was performed in order to show statistically significant improvement of the model (Table \ref{chisq-table}). \\ 
Performance can be compared in terms of WUPS as well. Wu-Palmer (WUP) Similarity can be used as an alternative to accuracy \cite{wu1994verbs}. \cite{malinowski2014multi} extended WUP similarity to the VQA task evaluation. WUP is based on how  the predicted answer semantically matches the ground truth. Given a predicted answer and a ground truth answer, WUPS computes a value between 0 and 1 based on their similarity. It computes similarity by considering the depths of the two synsets in WordNet, along with the depth of the  LCS  (Longest  Common Subsumer). WUPS is computed based on WUP. Given \textit{N} number of samples with \textit{A} being the ground truth answers and \textit{T} predicted answers, the formula is as follows: 
\begin{multline}
WUPS(A,T) = \frac{1}{N} \sum_{i=1}^{N} \min \\
\{ \prod_{a \in A^i}\max_{t \in T^i} WUP(a,t) , \prod_{t \in T^i}\max_{a \in A^i} WUP(a,t) \} . 100
\end{multline}
Here are some examples of the pure WUP score to give intuitions about the range: \textit{WUP(outside, outdoors)} = $0.92$, \textit{WUP(man,woman)} = $0.07$, \textit{WUP(land,earth)} = $1.0$ \textit{WUP(tree,water)}= $0.14$ and \textit{WUP(dog,wolf)} = $0.93$.\\ 

"WUPS at $0.9$" applies a threshold and considers a predicted answer correct if the WUPS score is higher than $0.9$.
"WUPS at $1.0$" corresponds to accuracy and \cite{malinowski2014multi} found that for VQA tasks a WUP score of around $0.9$ is required for precise answers. Table ~\ref{perfromance-table} shows performance in terms of "WUPS at $0.9$". The improvement based on WUPS, using multi-task approach, is almost similar to that based on accuracy. \\
The accuracy of frame element classification is initially $90\%$ and gets up to $99.68\%$ at the end of the training. The performance on test data is $99.32\%$. This improvement,as we will discuss later, helps the model to provide more consistent responses and to regularize the model. \\

\begin{center}
\begin{figure}[h!]
\includegraphics[width=0.5\textwidth]{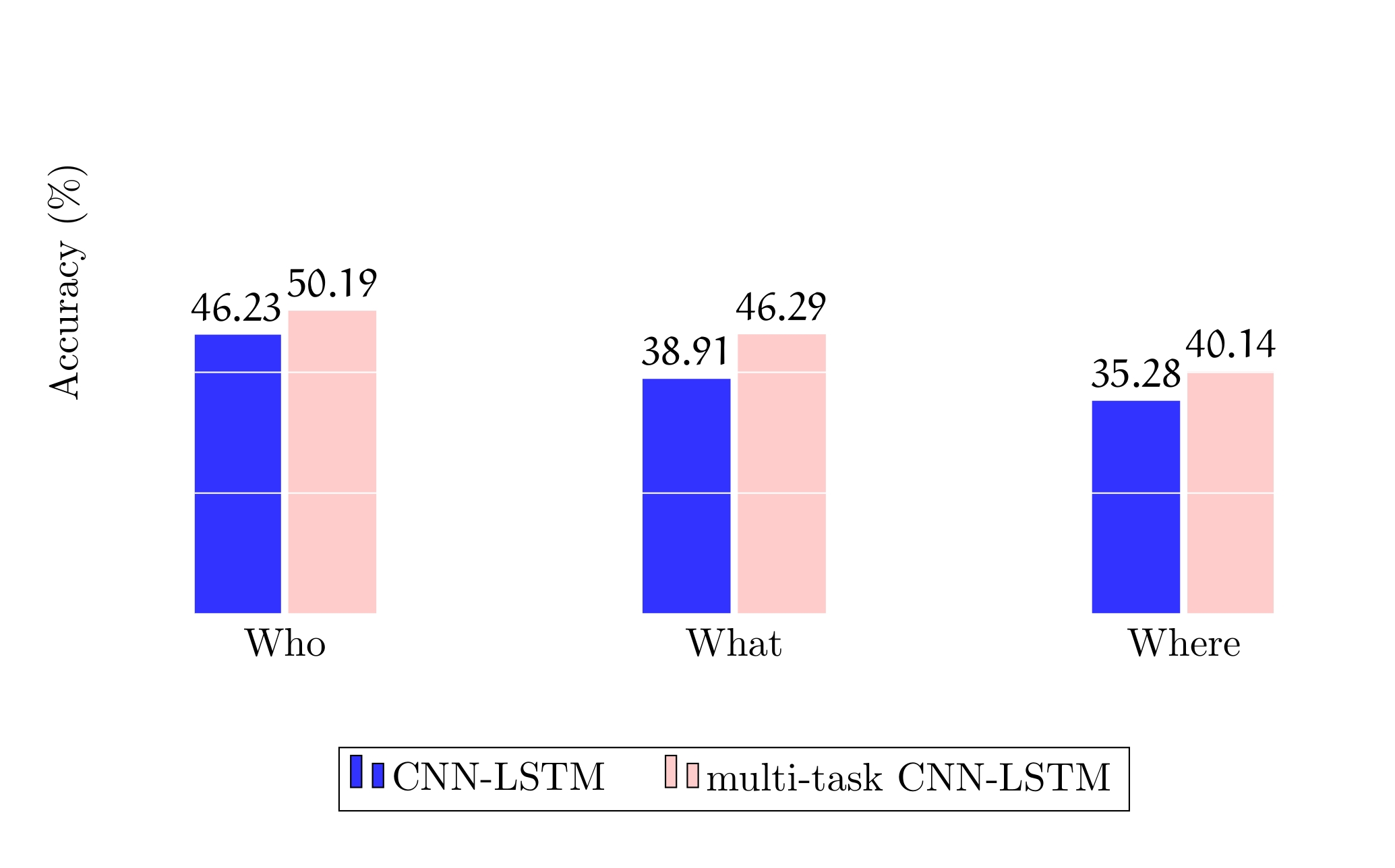}
\caption{Evaluation by wh-question type of the question}
\label{fig:perq-eval}
\end{figure}
\end{center}

\begin{table}[h!]
\begin{center}
\resizebox{0.4\textwidth}{!}{%
\begin{tabular}{|c|c|c|} 
\cline{2-3}
\multicolumn{1}{c|}{} & Correct & Incorrect  \\
\hline 
CNN-LSTM  & 34905  & 53065\\
\hline 
multi-task CNN-LSTM  & 39522 & 48448\\

\hline 
\end{tabular}}%
\end{center}
\caption{The chi-square statistic is $496.1854$. The p-value is $<0.01$ and the result is significant.}
\label{chisq-table}
\end{table}

\begin{center}
\begin{figure*}[h!]
\includegraphics[width=0.99\textwidth]{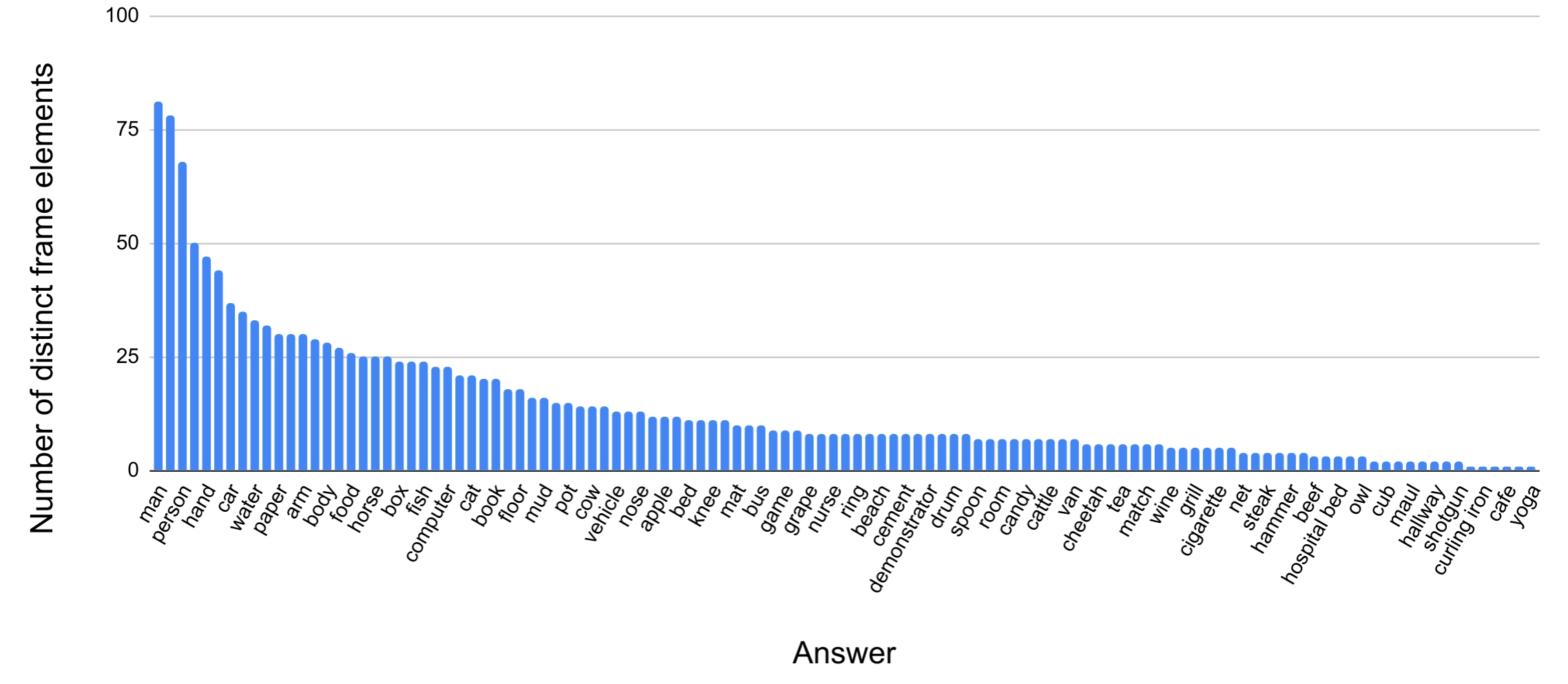}
\caption{Distinct frame element frequency for different answers.}
\label{fig:answerrolefrequency}
\end{figure*}
\end{center}

\textbf{Frame element classification:} The hyper-class augmentation model utilizes frame element classification for better representation learning of the VQA task. As discussed earlier, the accuracy of the frame element classification is $99.32\%$. One important reason for such high performance is the  frame element dependency on the input question while it is independent of the input image. For example for the question \textit{"who is cooking ?"} the frame element is always \textit{AGENT} for all images about cooking. This results in a huge amount of data to train the frame element classification resulting in almost perfect performance. \\
It is interesting to know how frame element classification affects the predicted answer and how consistent it is with the correct answer and predicted answer. We consider the correct or predicted answer to be consistent with the frame element if there is at least one training sample labeled with both the answer and the frame element. For example \textit{$<$bear, AGENT$>$} and \textit{$<$bear, CHASEE$>$} are consistent but \textit{$<$bear, PLACE$>$} and \textit{$<$bear, TOOL$>$} are inconsistent. Figure \ref{fig:answerrolefrequency} shows the frequency of distinct frame elements for a subset of answers. For example \textit{man, car, telephone, bear} and \textit{cafe} are fillers of $81$, $37$, $20$, $8$ and $1$ distinct frame elements in the training samples respectively. An answer is consistent with the set of distinct frame elements it fills and inconsistent with others. \\
The almost perfect accuracy of the frame element classifier confirms its output is almost always consistent with the correct answer. Now the question is, how much does frame element classification help the predicted answer to be consistent with the semantic frame? Employing the consistency criterion, the consistency of the CNN-LSTM model is $97.56\%$ and multi-task CNN-LSTM $99.94\%$. This shows a $2.38\%$ improvement. In other words, augmenting the frame element classification decreases inconsistency in providing final responses. Consequently, the end-user would get more reasonable answers from the system.\\
\textbf{Fine-grained evaluation.} In order to perform a fine-grained analysis of the results, performance per question, per verb and per role are computed. Figure ~\ref{fig:perq-eval} shows a performance comparison based on the  \textit{wh-question type} of the question. Multi-task CNN-LSTM performs better for who (4\%), what (8\%) and where (5\%) when compared to CNN-LSTM. Exploring performance per verb, we can see for example \textit{cooking} improves from 30.12\% to 44.58\% and \textit{buying} from 27.42\% to 64.52\%. Exploring performance per role, for example, the multi-task approach improves \textit{AGENT} from 48.78\% to 52.29\%, \textit{PLACE} from 34.75\% to 39.52\%  and \textit{ITEM} from 32.27\% to 39.65\%. Table \ref{per-vr-table} shows a different view of the performance difference between CNN-LSTM and the multi-task version. About 55\% of verbs improve by less than 10\%. \textit{whipping, buying, sketching, scooping, making} improve by more than 30\%. \textit{spanking, ejecting, farming, hitting, harvesting, moistening} decline by more than 15\%.

\begin{table}[h!]
\begin{center}
\resizebox{0.3\textwidth}{!}{%
\begin{tabular}{c c c} 
\hline
Accuracy & Verb & Role \\
Difference & Frequency & Frequency \\
Range   & & \\
\hline 
(-40\%,-30\%] & & 3 \\
(-30\%,-20\%] & 2 & 3\\
(-20\%,-10\%] & 10 & 5\\
(-10\%,0\%) & 67 & 24\\
0\% & 27 & 32\\
(0\%,10\%] & 269  & 68\\
(10\%,20\%] & 100 & 24\\
(20\%,30\%] & 15 & 13\\
(30\%,40\%] & 6 & \\
(40\%,50\%] & & 4\\
(50\%,60\%] & & 2\\
 ... & & \\ 
100\% & & 1\\
\hline 

\end{tabular}}%
\end{center}
\caption{Performance evaluation grouped by performance intervals showing verb frequency and role frequency in each group.}
\label{per-vr-table}
\end{table}

\section{Conclusions}
In this paper, we explained how we used the imSitu annotations to build a VQA dataset with semantic verb information. Furthermore, we proposed a multitask learning approach in order to augment a CNN-LSTM VQA model. The approach boosts performance and shows the benefit of using verb semantics in answering questions about images. The proposed model utilizes semantic frame elements in order to answer the input question about the image. We evaluated the proposed model showing $5\%$ improvement in \textit{accuracy} and "\textit{WUPS at $0.9$}". \\
We provided a justification of why the proposed hyper-class augmentation idea works and also explored its effect through different analysis. However additional theoretical analysis would enrich the proposed VQA model and system. One hypothesis would be to consider the semantic information equivalent to context or context-aware information \cite{sli2016bandit} \cite{gentile2017context}. \\
In this work we created a VQA dataset where questions are annotated with precise frame element information. Another approach would be to employ a semantic role labeling model in order to approximately extract frame element information for any question of an available VQA dataset, and  then explore how frame element augmentation would work. \\
The hyper-class augmentation is a novel technique in the context of VQA. This idea can be generalized by augmenting the VQA models with answer types, task types, and other auxiliary information by means of the multi-task learning paradigm.\\

\bibliographystyle{./IEEEtran}
\bibliography{./IEEEabrv,./ieee-icsc-2020}

\end{document}